\documentclass[conference]{IEEE_ref}
\IEEEoverridecommandlockouts
\usepackage{cite}
\usepackage{amsmath,amssymb,amsfonts}
\usepackage{algorithmic}
\usepackage{graphicx}
\usepackage{textcomp}
\usepackage{tabularx}
\usepackage{xcolor}
\usepackage{multirow}
\usepackage{adjustbox}
\usepackage{tabularx}
\usepackage{array}
\usepackage{colortbl}
\usepackage{booktabs}
\usepackage{stfloats}
\usepackage{lineno,hyperref}
\hypersetup{hidelinks,
	colorlinks=true,
	allcolors=black,
	pdfstartview=Fit,
	breaklinks=true}
\def\BibTeX{{\rm B\kern-.05em{\sc i\kern-.025em b}\kern-.08em
    T\kern-.1667em\lower.7ex\hbox{E}\kern-.125emX}}
\begin{document}

\title{Test-time Alignment-Enhanced Adapter for Vision-Language Models\\

}

\author{\IEEEauthorblockN{1\textsuperscript{st} Baoshun Tong}
\IEEEauthorblockA{\textit{school of computer science and engineering} \\
\textit{sun yat-sen university}\\
Guangzhou, China \\
tongbsh@mail2.sysu.edu.cn}
\and
\IEEEauthorblockN{2\textsuperscript{nd} Kaiyu Song}
\IEEEauthorblockA{\textit{school of artificial intelligence} \\
\textit{sun yat-sen university}\\
Guangzhou, China \\
songky7@mail2.sysu.edu.cn}
\and
\IEEEauthorblockN{3\textsuperscript{rd} Hanjiang Lai$^{\ast}$}
\IEEEauthorblockA{\textit{school of computer science and engineering} \\
\textit{sun yat-sen university}\\
Guangzhou, China \\
laihanj3@mail.sysu.edu.cn}

}

\maketitle

\begin{abstract}
Test-time adaptation with pre-trained vision-language models (VLMs) has attracted increasing attention for tackling the issue of distribution shift during the test phase. 
While prior methods have shown effectiveness in addressing distribution shift by adjusting classification logits, they are not optimal due to keeping text features unchanged. To address this issue, we introduce a new approach called Test-time Alignment-Enhanced Adapter (TAEA), which trains an adapter with test samples to adjust text features during the test phase. We can enhance the text-to-image alignment prediction by utilizing an adapter to adapt text features. Furthermore, we also propose to adopt the negative cache from TDA as enhancement module, which further improves the performance of TAEA. Our approach outperforms the state-of-the-art TTA method of pre-trained VLMs by an average of 0.75\% on the out-of-distribution benchmark and 2.5\% on the cross-domain benchmark, with an acceptable training time. Code will be available at \href{https://github.com/BaoshunWq/clip_TAEA}{https://github.com/BaoshunWq/clip-TAEA}.
\end{abstract}


\section{INTRODUCTION}
\label{sec:intro}

Recently, test-time adaptation (TTA)~\cite{wang2020tent}, which adapts the model to unlabeled test data, has drawn much interest due to the distribution shift problem between the training and testing data.
Since foundational vision-language models such as CLIP (Contrastive Language-Image Pretraining)~\cite{radford2021learning} have achieved excellent results in many downstream tasks~\cite{ming2022delving,zhou2023zegclip}, some researchers~\cite{shu2022test,feng2023diverse,karmanov2024efficient} have pay attention to the pre-trained vision-language models for TTA. Test-time Prompt Tuning (TPT)~\cite{shu2022test} firstly addressed the TTA issue by learning an adaptive prompt on the fly with test samples with the zero-shot generalization in VLMs. And DiffTPT~\cite{feng2023diverse} leveraged pre-trained diffusion models to generate diverse and informative new data to learn a better prompt. 
Later, Training-free dynamic adapter (TDA)~\cite{karmanov2024efficient}  is proposed to adapt CLIP to downstream tasks by building two dynamic caches without training. This type of test-time adaptation heavily relies on the vision-language alignment capability of CLIP.

Despite the efficiency and effectiveness achieved by prior methods, the distribution shift between test data and training data can also damage the alignment capability of pre-trained vision-language models since the hand-crafted text prompts are challenging to design for the unknown test distribution, thereby reducing classification accuracy.
Although the TPT and DiffTPT methods attempt to learn a prompt to adjust text features, its significant training overhead is not in line with the requirements of the testing phase \cite{karmanov2024efficient}. In summary, we contemplate the following question: Can we efficiently adjust text features to further enhance the capability of text-image alignment under the unsupervised condition of TTA?

In this paper, we propose a novel Test-time Alignment-Enhanced Adapter (TAEA) to efficiently and effectively enhance the capability of text-image alignment.
Our method consists of two modules. The first is the adapter module, which introduces a lightweight attention \cite{vaswani2017attention} block to adapt text category embedding according to the downstream images during the test phase. We use the original text features as a query while employing the test image features as key and value. Then, we train a gated single-head attention block to bridge the gap \cite{nam2021reducing} between test-time image features and text features for each category. With the help of the test-time knowledge, this process can be considered as a learnable module to match the text features with test samples. The second one is the enhancement module, which helps to mitigate the prediction errors of high entropy or bias to certain predictions due to training with pseudo labels and further enhances the performance of our adapter.

Through experiments on the out-of-distribution (OOD) benchmark and cross-domain benchmark \cite{shu2022test}, we observe that our TAEA can outperform the existing state-of-the-art test-time adaptation methods. The contributions of our study are summarized as follows: 
1) We propose test-time alignment-enhanced adapter (TAEA), a simple yet effective test-time adaptation method for vision-language models to help adapt text features to better align with test image features efficiently.
2) Under the unsupervised condition of TTA, we mitigate the problem of poor alignment capability of pre-trained vision-language models on the test dataset caused by distribution shift.
3) Experimental results show that our TAEA can outperform existing state-of-the-art test-time adaptation methods of pre-trained vision-language models.

\section{RELATEDWORK}
\label{sec:format}

% \subsection{Test-time adaptation of VLMs.}
% Recent TTA methods can be divided into two categories: conventional TTA methods and TTA methods specifically designed for vision-language models. 
% Conventional TTA methods mainly utilize each batch of testing samples to update partial weights \cite{iwasawa2021test,sun2020test}, normalization statistics \cite{schneider2020improving}, or a combination of both \cite{wang2020tent}. However, for vision-language models, such methods may not be applicable. 
% To adapt VLMs during the test phase, \cite{shu2022test} firstly proposes test-time prompt tuning (TPT), which can learn adaptive prompts on the fly with a single test sample. Furthermore, \cite{feng2023diverse} proposes DiffTPT, which leverages pre-trained diffusion models to generate diverse and informative new data to the extent of learning a better prompt. \cite{karmanov2024efficient} propose a training-free dynamic adapter (TDA), which constructs two caches to adjust the logits by calculating the similarity between the test image and the images in the cache. 

% \subsection{Adapter.}
Adapter-based architectures have gained attention for efficiently incorporating task-specific modifications into pre-trained models. Reference \cite{gao2024clip} proposes CLIP-Adapter, which adopts an additional bottleneck layer to learn new features and performs residual-style feature blending with the original pre-trained features. Reference \cite{zhang2022tip} proposes Tip-Adapter, which constructs the adapter through a key-value cache model from the few-shot training set and updates the prior knowledge encoded in CLIP through feature retrieval. Reference \cite{song2023meta} proposes Meta-Adapter, which constructs a lightweight network based on the gated multi-head attention ~\cite{vaswani2017attention} mechanism, to bridge the gap between few-shot image features and text features for each category. Reference \cite{Chen_2023_sam_adapter} proposes SAM-Adapter, which incorporates domain-specific information or visual prompts into the segmentation network via using simple yet effective adapters. Reference \cite{mou2024t2i_adapter} proposes to learn low-cost T2I-Adapters to align internal knowledge in text-to-image (T2I) models with external control signals, while freezing the original large T2I models. And~\cite{he2021acl} states that the adapter-based tuning better mitigates forgetting issues than fine-tuning since it yields representations with less deviation from those generated by the initial pre-trained language model.

\section{METHOD}
\label{sec:method}

\subsection{Reviews of CLIP and TDA}

Given a test image $x_t$, a global visual feature $f_{test} = E_i(x_t)$ is obtained by CLIP's visual encoder $E_i$. Similarly, the corresponding text features $\omega$ can be encoded using CLIP's text encoder $E_t$. The text is composed of hand-crafted templates, one typical form being “a photo of [CLASS]” and a specific category name, such as “dog.” As a result, CLIP's prediction $y$ for a test image is obtained by the matching score as follows:
\begin{equation}
        P_{\mathrm{clip}}(f_{\mathrm{test}})=f_\text{test}\mathbf{\omega}^T.
        \label{eq:1-zero clip}
\end{equation}
To better adapt CLIP to TTA tasks, TDA \cite{karmanov2024efficient} comprises two dynamic key-value caches, each storing a dynamic queue of few-shot test features as keys and their corresponding pseudo-labels as values. The first one is positive cache, which aims to collect high-quality few-shot pseudo labels \textbf{$\hat{L_p}$} as positive values and the corresponding features \textbf{$Q_p$} as keys. The second one is the negative cache,  which aims to gather CLIP-generated image features to \textbf{$Q_n$}, and the corresponding negative pseudo labels to \textbf{$\hat{L_n}$}.
Finally, the prediction of TDA can be formulated by combining the negative cache, the positive cache, and the pre-trained CLIP model as follows:
\begin{equation}
    P_{\mathrm{TDA}}(f_{\mathrm{test}})=P_{\mathrm{clip}}(f_{\mathrm{test}})+P_{\mathrm{pos}}(f_{\mathrm{test}})+P_{\mathrm{neg}}(f_{\mathrm{test}}).
    \label{eq:4-TDA}
\end{equation}

Although TDA has achieved great efficiency and effectiveness, solely adjusting classification logits based on the dynamic cache while keeping text features unchanged is not optimal.

\begin{figure}
        
        \includegraphics[width=\linewidth]{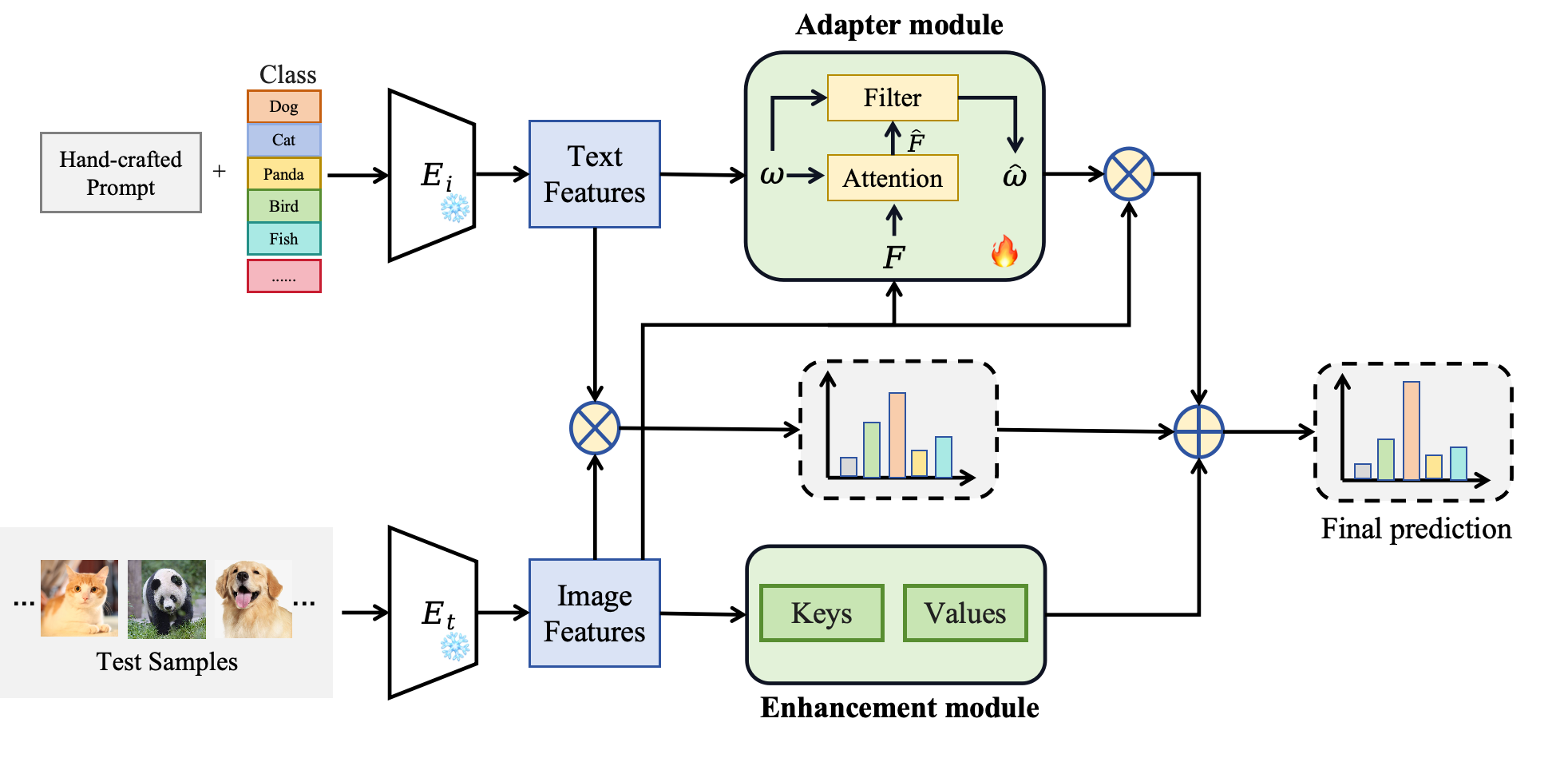}
        \caption{Overall pipeline of the proposed method. The adapter module aims to adjust hand-crafted text features and improve the capability of text-to-image alignment. The enhancement module aims to enhance the performance further.}
        \label{fig-pipeline}
\end{figure}

\begin{table*}[t]
\caption{\textbf{Results on the Cross-Domain Benchmark.} Same as the prior methods, the evaluation metric Average is calculated using the mean accuracy across all ten datasets.}
\label{cd-result}
\scriptsize
\setlength{\tabcolsep}{3.14mm}
\begin{tabular*}{\linewidth}{l*{11}{c}}
\toprule
Method & Aircraft & Caltech101 & Cars & DTD & EuroSAT & Flower102 & Food101 & Pets & SUN397 & UCF101 & Average \\
\midrule
CLIP-ResNet-50 & 16.11 & 87.26 & 55.89 & 40.37 & 25.79 & 62.77 & 74.82 & 82.97 & 60.85 & 59.48 & 56.63 \\
\hline

CoOp & 15.12 & 86.53 & 55.32 & 37.29 & 26.20 & 61.55 & 75.59 & 87.00 & 58.15 & 59.05 & 56.18 \\
CoCoOp & 14.61 & 87.38 & 56.22 & 38.53 & 28.73 & 65.57 & 76.20 & 88.39 & 59.61 & 57.10 & 57.23 \\
\hline

TPT & 17.58 & 87.02 & 58.46 & 40.84 & 28.33 & 62.69 & 74.88 & 84.49 & 61.46 & 60.82 & 57.66 \\
DiffTPT & 17.60 & 86.89 & \textbf{60.71} & 40.72 & 41.04 & 63.53 & \textbf{79.21} & 83.40 & 62.72 & 62.67 & 59.85 \\
TDA & 17.61 & \textbf{89.70} & 57.78 & 43.74 & 42.11 & \textbf{68.74} & 77.75 & 86.18 & 62.53 & \textbf{64.18} & 61.03 \\
\rowcolor{gray!30}
TAEA (ours) & \textbf{18.60} & 87.87 & 57.82 & \textbf{51.77} & \textbf{59.05} & 67.24 & 78.44 & \textbf{88.01} & \textbf{66.44} & 61.62 & \textbf{63.69} \\
\midrule
CLIP-ViT-B/16 & 23.22 & 93.55 & 66.11 & 45.04 & 50.42 & 66.99 & 82.86 & 86.92 & 65.63 & 65.16 & 64.59 \\
\hline

CoOp & 18.47 & 93.70 & 64.51 & 41.92 & 46.39 & 68.71 & 85.30 & 89.14 & 64.15 & 66.55 & 63.88 \\
CoCoOp & 22.29 & 93.79 & 64.90 & 45.45 & 39.23 & 70.85 & 83.97 & 90.46 & 66.89 & 68.44 & 64.63 \\
\hline

TPT & 24.78 & 94.16 & 66.87 & 47.75 & 42.44 & 68.98 & 84.67 & 87.79 & 65.50 & 68.04 & 65.10 \\
DiffTPT & 25.60 & 92.49 & 67.01 & 47.00 & 43.13 & 70.10 & \textbf{87.23} & 88.22 & 65.74 & 62.67 & 65.47 \\
TDA & 23.91 & \textbf{94.24} & \textbf{67.28} & 47.40 & 58.00 & \textbf{71.42} & 86.14 & 88.63 & 67.62 & \textbf{70.66} & 67.53 \\
\rowcolor{gray!30}
TAEA (ours) & \textbf{27.45} & 93.75 & 66.53 & \textbf{52.66} & \textbf{68.74} & 71.21 & 86.70 & \textbf{91.69} & \textbf{71.10} & 70.47 & \textbf{70.03} \\
\bottomrule
\end{tabular*}
\end{table*}

\subsection{Our method}
This section presents a new approach called Test-time Alignment-Enhanced Adapter (TAEA) to adjust text features. And we employ an enhancement module to further enhance text-image alignment capability under the unsupervised condition of TTA. Fig.~\ref{fig-pipeline} shows an overview of our proposed method.

\textbf{Adapter module.} The adapter module introduces two steps. The first step involves retrieving test knowledge from the test image with attention mechanisms. This can obtain image features related to text features. The second step employs a gating network to aggregate hand-crafted text features with the image features obtained in the first step. After completing the above two steps, we can obtain the text features aligned with the image.

Specifically, given the test images, we can obtain the image features $F$ and the hand-crafted text features $\omega$ with the encoder of CLIP. The text features $\omega$ is composed of hand-crafted prompt and $N$-class names. Then, we match the test image features with text features using a single attention block. Thanks to the attention~\cite{vaswani2017attention} mechanism, we can get better image features related to the text features by treating $\omega$ as the query and employing the test image features $F$ as both key and value.
To reduce training overhead and achieve a lightweight adapter, following the previous works~\cite{song2023meta}, we leverage MLP as the attention mechanism, which could be defined as:
\begin{equation}
    \hat{\mathbf{F}}=F^\top\sigma((\omega W_W^\top)(F W_F^\top)^\top/\sqrt{D}),
    \label{eq:5-attention}
\end{equation}
where $\hat{\mathbf{F}}$ represents the image feature related to text feature. The $W_F$ and $W_W$ indicate the weights of MLP layers. The $\sigma$ denotes the Softmax function and $D$ is the scaling factor.

After obtaining the $\hat{\mathbf{F}}$, we introduce a learnable gate block $f(\cdot)$ to match $\hat{\mathbf{F}}$ and text features $\omega$, where $f(\cdot)$ only contains a single MLP layer. In this way, we can generate a modulation scalar to filter knowledge and control the ratio between text and image features in the test time~\cite{song2023meta}. Finally, we could fine-tune the text feature as follows:
\begin{equation}
    \hat\omega=\omega+f(\omega)\odot\hat{\mathbf{F}},
    \label{eq:6-filter}
\end{equation}
where $\odot$ denotes Hadamard product. After training the gate block, $f(\cdot)$ can adjust the match ratio according to the $\omega$ and address distribution shift.
The adapter could be defined as follows: 
\begin{equation}
    P_{\mathrm{adapter}}(f_{\mathrm{test}})=\gamma\frac{\hat\omega^\top f_{test}}{\|\hat\omega\|\|f_{test}\|},
    \label{eq:7-adapter}
\end{equation}
where $f_{test}$ is the test image feature obtained by encoder and $\gamma$ is a hyper-parameter. 

To train our adapter module, we select samples with lower entropy to generate a pseudo label $L_{a}$, a one-hot encoded vector of a categorical distribution, and store them during the testing period of the first $\lambda*N$ samples. The $N$ represents the total number of images in the test set, and $\lambda$ is a hyperparameter to determine the timing of training the adapter. After the model has tested $\lambda*N$ samples, we train the adapter module to update text features with these test samples. We use cross-entropy as the loss function.

The effectiveness of adjusting text features is theoretically guaranteed by non-local filters~\cite{wang2018non,yan2020deep,cao2019gcnet}. Intuitively, similar to the non-local filters, \cite{song2023meta} states that an adapter based on gated attention could disregard some outlier samples while paying more attention to the samples related to the category description, resulting in robust feature representations. 

\textbf{Enhancement module.} To mitigate the risk of prediction errors due to high entropy or being biased to certain predictions, we propose to adopt the negative cache from TDA~\cite{karmanov2024efficient} as the plug-and-play module for further enhancing the performance of our adapter. As stated by ~\cite{karmanov2024efficient}, the negative cache is designed for negative learning. It aims to mitigate the negative impact of noisy pseudo-labels by introducing negative pseudo-labeling to identify class absence rather than presence.

In summary, with the help of the adapter and enhancement modules, our method can adjust hand-crafted text features to enhance the capability of text-image alignment. The overall formulation for computing classification results in our method is as follows:
\begin{equation}
    P_{\mathrm{TAEA}}(f_{\mathrm{test}})=P_{\mathrm{clip}}(f_{\mathrm{test}}) + P_{\mathrm{adapter}}(f_{\mathrm{test}}) + P_{\mathrm{neg}}(f_{\mathrm{test}}).
    \label{eq:8-TCA}
\end{equation}

\section{EXPERIMENT}

\subsection{Experimental Detail}

\textbf{Datasets.} Consistent with prior works~\cite{feng2023diverse,loshchilovdecoupled2017adaw}, we conduct main experiments on two benchmarks: out-of-distribution (OOD) benchmark and cross-domain benchmark. To assess the robustness and generalization, OOD benchmark consists of one ID dataset ImageNet~\cite{deng2009imagenet} and 4 out-of-distribution datasets derived from ImageNet: ImageNet-A \cite{hendrycks2021imagenet-a}, ImageNet-V2 \cite{recht2019imagenetv2}, ImageNet-R \cite{hendrycks2022imagenet-r}, and ImageNet-S \cite{wang2019imagenet-s}. On the other hand, to evaluate the adaptation ability during testing on datasets with different domain distributions, the cross-domain benchmark consists of 10 diverse image classification datasets, each from a distinct domain with different classes: Pets \cite{parkhi2012pets}, EuroSAT \cite{helber2019eurosat}, Aircraft \cite{maji2013fgvc}, Caltech101 \cite{fei2004caltech101}, UCF101 \cite{soomro2012ucf101}, Cars \cite{krause2013cars}, DTD \cite{cimpoi2014describing}, Flower102 \cite{nilsback2008flowers}, Food101 \cite{bossard2014food101}, and SUN397 \cite{xiao2010sun}.

\textbf{Baselines.} To evaluate the effectiveness of our method, we compare it with zero-shot methods, train-time adaptation methods, and test-time adaptation methods, including the recent state-of-the-art method. For the zero-shot method, We compared with public CLIP \cite{radford2021learning} results under ResNet-50~\cite{he2016resnet} and ViT-B/16~\cite{dosovitskiy2020vit}. For the train-time adaptation methods, we compared with CoOp \cite{zhou2022learning}, CoCoOp \cite{zhou2022conditional}, and Tip-Adapter \cite{zhang2022tip}. 
For the test-time adaptation methods, we compared with TPT \cite{shu2022test}, and its improved version, DiffTPT \cite{feng2023diverse}. We also compared against the challenging state-of-the-art method in this field: TDA \cite{karmanov2024efficient}. All results of the methods compared in the table are obtained from the~\cite{karmanov2024efficient} paper. 

\textbf{Implementation Details.} Same as prior works \cite{shu2022test,feng2023diverse,karmanov2024efficient}, we conducted experiments on out-of-distribution benchmark and cross-domain benchmark separately using ResNet-50 and ViT-B/16 backbones. Specifically, we use a batch size of 1, which means the test time adaptation is set for single-image scenarios. We set all database-related hyperparameters to be consistent with TDA. And we set $\lambda=0.25$, $\gamma=0.6$. For the training of the adapter module, We optimize the gated attention block with a batch size of 3 and use AdamW~\cite{loshchilovdecoupled2017adaw} optimizer with a learning rate of 0.001 and a cosine scheduler for 3 epochs. In all experiments, we evaluate a single NVIDIA Quadro RTX 6000 GPU.

\subsection{Main results}

\textbf{Results on the OOD benchmark.} The performances on the OOD benchmark are summarized in Table ~\ref{ood-result}. Our method has surpassed the method that adjusts text features similarly and the current state-of-the-art method on average. Specifically, compared to the TPT method, which adjusts text features similarly, our method outperforms TPT on both ResNet-50 and ViT-B/16 architectures, improving all accuracy by 2.87\% and 3.32\% on average, respectively. Furthermore, compared to the current state-of-the-art method TDA, our method outperforms TDA on both ResNet-50 and ViT-B/16 architectures, improving all accuracy by 0.55\% and 0.75\% on average, respectively.

As shown in Table~\ref{testing_time_accuracy_gain}, to further evaluate the efficiency and effectiveness of our method, we conducted experiments on ImageNet validation dataset to test both accuracy and testing time. 
When compared to the current state-of-the-art method (TDA), although our approach requires additional testing time (requiring an additional 2min) due to minimal training requirements, we have achieved substantial improvements in accuracy (+3.82\%). 
Compared to methods requiring similar training for adjusting text features, our approach dramatically reduces the testing time from 12h 50min by TPT and even more from 34h 45min by DiffTPT, down to just 18 minutes.

\begin{table}
\centering
\caption{\textbf{Results on the OOD Benchmark.} To demonstrate the effectiveness of our method, we conducted a comparison of two versions of the CLIP backbone: ResNet-50, and ViT-B/16.}
\label{ood-result}
\scriptsize
\setlength{\tabcolsep}{0.55mm}
\begin{tabularx}{\linewidth}{cccccccc}
\toprule
\text{Method} & \text{ImageNet} & \text{ImageNet-A} & \text{ImageNet-V2} & \text{ImageNet-R} & \text{ImageNet-S} & \text{Average}  \\
\midrule 
\text{ResNet-50} & 59.81 & 23.24 & 52.91 & 60.72 & 35.48 & 46.43  \\
\hline
\text{CoOp} & 63.33 & 23.06 & 55.40 & 56.60 & 34.67 & 46.61  \\
\text{CoCoOp} & 62.81 & 23.32 & 55.72 & 57.74 & 34.48 & 46.81  \\
\text{Tip-Adapter} & 62.03 & 23.13 & 53.97 & 60.35 & 35.74 & 47.04  \\
\hline
\text{TPT} & 60.74 & 26.67 & 54.70 & 59.11 & 35.09 & 47.26  \\
\text{DiffTPT} & 60.80 & \textbf{31.06} & 55.80 & 58.80 & 37.10 & 48.71  \\
\text{TDA} & 61.35 & 30.29 & 55.54 & \textbf{62.58} & \textbf{38.12} & 49.58  \\
\rowcolor{gray!30}
\text{TAEA (ours)} & \textbf{63.63} & 30.72 & \textbf{56.15} & 62.46 & 37.70 & \textbf{50.13}  \\

\midrule
\text{ViT-B/16} & 68.34 & 49.89 & 61.88 & 77.65 & 48.24 & 61.20  \\
\hline
\text{CoOp} & 71.51 & 49.71 & 64.20 & 75.21 & 47.99 & 61.72  \\
\text{CoCoOp} & 71.02 & 50.63 & 64.07 & 76.18 & 48.75 & 62.13  \\
\text{Tip-Adapter} & 70.75 & 51.04 & 63.41 & 77.76 & 48.88 & 62.37  \\
\hline
\text{TPT} & 68.98 & 54.77 & 63.45 & 77.06 & 47.94 & 62.44  \\
\text{DiffTPT} & 70.30 & 55.68 & 65.10 & 75.00 & 46.80 & 62.28  \\
\text{TDA} & 69.51 & \textbf{60.11} & 64.67 & 80.24 & 50.54 & 65.01  \\
\rowcolor{gray!30}
\text{TAEA (ours)} & \textbf{71.71} & 60.10 & \textbf{65.23} & \textbf{80.73} & \textbf{51.04} & \textbf{65.76}  \\
\bottomrule
\end{tabularx}
\end{table}

\begin{table}[bp]

\caption{\textbf{Comparisons of efficiency (Testing Time) and effectiveness (Accuracy).} The last column in the table represents the accuracy gain relative to the baseline clip.} 
\label{testing_time_accuracy_gain}
\setlength{\tabcolsep}{4mm}{
\begin{tabularx}{\linewidth}{p{2.4cm} *{7}{c}}
\hline
Method & Testing Time & Accuracy & Gain \\
\hline
ResNet-50 & \underline{\textbf{12 min}} & 59.81 & 0 \\
TPT & 12h 50min & 60.74 & +0.93 \\
DiffTPT & 34h 45min & 60.80 & +0.99 \\
TDA  & 16 min & 61.35 & +1.54 \\
\rowcolor{gray!30}
TAEA (ours) & 18 min & \textbf{63.63} & \textbf{+3.82} \\
\hline
\end{tabularx}
}
\end{table}

\textbf{Results on the cross-domain benchmark.} The performances on the cross-domain benchmark are summarized in Table~\ref{cd-result}. 
Our method not only surpasses the methods (i.e., TPT, DiffTPT) that adjust text features similarly but also exceeds the current state-of-the-art method TDA on average, reaching a new state-of-the-art. Specifically, compared to the TPT method, which adjusts text features similarly, our method outperforms TPT on both ResNet-50 and ViT-B/16 architectures, improving all accuracy by 6.03\% and 4.93\% on average, respectively. Furthermore, compared to the current state-of-the-art method TDA, our method outperforms TDA on both ResNet-50 and ViT-B/16 architectures, improving all accuracy by 2.66\% and 2.5\% on average, respectively.

\subsection{Ablation Study}
\begin{figure}

\begin{minipage}[b]{.48\linewidth}
  \centering
  \centerline{\includegraphics[width=4.0cm]{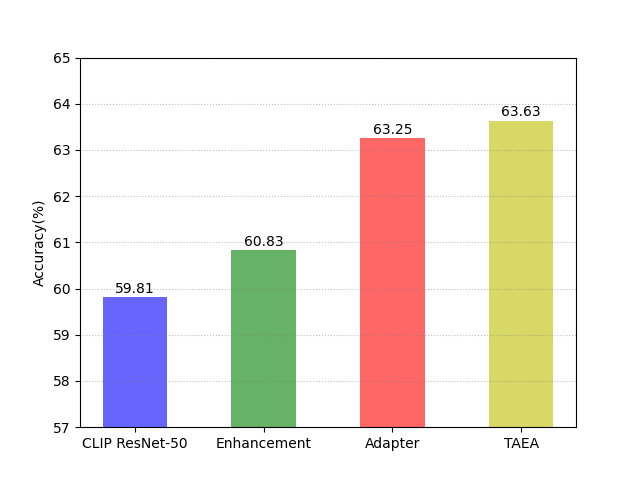}}
%  \vspace{1.5cm}
  \centerline{(a) Results on Imagenet.}\medskip
\end{minipage}
\hfill
\begin{minipage}[b]{0.48\linewidth}
  \centering
  \centerline{\includegraphics[width=4.0cm]{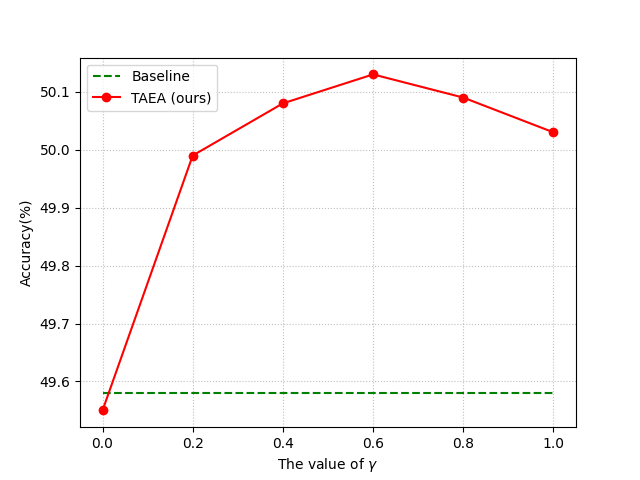}}
%  \vspace{1.5cm}
  \centerline{(b) Results on OOD benchmark.}\medskip
\end{minipage}
\caption{The results of ablation study.}
\label{fig:ablation}
\end{figure}

To validate the effectiveness of our proposed method, we conducted ablation experiments on the ImageNet~\cite{deng2009imagenet}. TAEA consists of the adapter module and the enhancement module from negative cache \cite{karmanov2024efficient}. 
We first evaluated the effectiveness of using a single enhancement module and a single adapter module. As shown in Fig.~\ref{fig:ablation}(a), each module can outperform the original CLIP-ResNet-50. Our adapter module exceeds CLIP and surpasses enhancement module, demonstrating its ability to effectively adjust text features and enhance the alignment capability between text and image. Furthermore, combining the enhancement module with the adapter can further improve performance on ImageNet, which is our TAEA method.

Furthermore, we also conduct ablation studies on $\gamma$ in (\ref{eq:7-adapter}). $\gamma$ is a parameter used to control the weighting of the adjusted text features for classification. Specifically, We experimented with a challenging OOD benchmark using $\gamma=0$, 0.2, 0.4, 0.6, 0.8, 1.0. It's important to note that $\gamma=0$ indicates that the model did not use the adapter to adjust the text, which means that the text features remain unchanged during the testing phase. In Fig.~\ref{fig:ablation}(b), we summarize the average results. The result of baseline is from~\cite{karmanov2024efficient}

\section{CONCLUSION} 
To improve the alignment capability between text and image during the testing phase, we present a novel test-time adaptation approach TAEA called the Test-time Alignment-Enhanced Adapter, which integrates test distribution knowledge and adapts text features with a gated attention mechanism.
In terms of performance, the results of extensive experiments demonstrate that TAEA outperforms state-of-the-art test-time adaptation methods. In terms of efficiency, we have significantly reduced the testing time compared to previous methods that adjust text features by learning prompts.

\vfill\pagebreak

\bibliographystyle{IEEE_icassp2025/IEEE_ref}
\bibliography{IEEE_icassp2025/IEEE_ref}

\end{document}